\documentclass[a4paper]{article}
\usepackage{acra}
\makeatletter
\let\normalsize\@normalsize
\makeatother
\normalsize
\usepackage[font=small]{caption}
\usepackage{amsmath,amssymb}
\usepackage{booktabs}
\usepackage{graphicx}
\usepackage{xcolor}
\usepackage{tikz}
\usetikzlibrary{arrows.meta,calc,fit,positioning}
\usepackage[hidelinks]{hyperref}
\usepackage{orcidlink}

\title{Linear Recurrent Memory Suffices to Distil a World-Model Policy for Robot Air Hockey}
\author{F. Olivia Fan\,\orcidlink{0009-0005-1607-3954}
\and Oliver Obst\,\orcidlink{0000-0002-8284-2062}\\
UNSW Sydney, NSW 2052, Australia}

\begin{document}
\maketitle

\begin{abstract}
Does memory-dependent control need nonlinear recurrent dynamics? We study simulated air-hockey defence under temporary loss of puck tracking. A DreamerV3 teacher outperforms a memoryless policy under tracking loss, and erasing its recurrent state at the moment tracking fails cuts its save rate sharply, so the task requires memory. We distil this teacher into compact recurrent policies with a 64-dimensional state, combining a diagonal linear recurrence with an optional rank-$k$ nonlinear innovation while the observation encoder and action head stay nonlinear. Across five matched seeds, the purely linear recurrent model ($k=0$) matches both the GRU baseline and the teacher throughout the tested range of tracking loss. Increasing nonlinear innovation rank provides no measured benefit. This result comes from a fresh test split that was opened only after all models and analyses were frozen. Under 5\,mm observation noise, without retraining, the linear student keeps a higher save rate than the GRU baseline, while using fewer recurrent parameters and less computation. These results suggest that, for this memory-dependent control task, nonlinear representation learning around a simple linear memory can be sufficient, and that nonlinear recurrent dynamics are not necessarily required.
\end{abstract}

\section{Introduction}

An air-hockey defender that loses sight of the puck must keep acting without knowing where the puck is or where it is going. In our task, a simulated arm defends its goal against a single launched puck whose observations are masked for up to 400\,ms. The problem is not specific to air hockey. Any vision-based controller faces it when occlusion, dropped frames or sensing faults interrupt tracking~\cite{liu2024retrospective}. Air hockey compresses the problem into a fast and exactly replayable form. While the puck is hidden, defence can only rely on information retained from earlier observations. Control runs at 50\,Hz, the rate set by the robot air-hockey challenge framework~\cite{liu2024retrospective} and applied by our teacher's training stack~\cite{orsula2024airhockey}. At that rate, a fixed-size recurrent state can carry the needed information without storing every past observation. In this work, we ask how complex that recurrent update needs to be. We answer in one controlled simulated air-hockey defence setting, with all students distilled from a single frozen teacher. Since everything else is held fixed, any difference between students comes from the recurrent update alone.

We study this question through policy distillation. We first train a DreamerV3 agent~\cite{hafner2023mastering} as the teacher. It learns a world model, so its actions depend on the whole history of observations, and it defends well through tracking loss. Smaller students then learn to act like the teacher, step by step and across whole episodes, without access to its latent state. Each student must therefore build its own compact summary of the history that control requires.

The physics during the blackout is simple. A hidden puck keeps sliding at nearly constant velocity, changing direction only when it bounces off a wall. A defender that remembers recent puck positions can therefore estimate the puck's velocity and project its path forward. This suggests that the useful work of memory is carrying information forward rather than transforming it. Can a student with a purely linear recurrent update behave like the teacher? To find out, we only vary the recurrent transition. The encoder, the action head, the 64-dimensional recurrent state, the training budget, the optimisation schedule and the paired evaluation all stay fixed. Our structured family adds a nonlinear innovation of rank $k$ on top of a diagonal linear recurrence, with $k\in\{0,1,2,4\}$. We compare these students with feed-forward, finite-history, and GRU-64 students.

Before comparing architectures, we establish that the task requires memory. We construct observation-alias shot pairs. In each pair, two different puck trajectories arrive at exactly the same observation at the moment the blackout begins, yet defending them requires different actions. A policy without memory must respond to both shots identically, so it must fail at least one of them. Only the history of the shot tells the two apart. We also test the teacher directly. At the moment tracking is lost, we erase its recurrent state and let it keep playing. The drop in its save rate measures how much of its defence rests on information carried over from the visible period.

This paper makes three contributions. First, a defence task under tracking loss, built around observation-alias pairs, with interventions on behaviour and on recurrent state that establish the task depends on memory. Second, a structured family of students in which a single rank parameter controls how much nonlinear recurrence the update has, while everything around it stays fixed. Third, a paired evaluation over five seeds in which the fully linear student matches the save rate of GRU-64 through 400\,ms of tracking loss under clean observations, using 88\% fewer parameters in its recurrent core. Nonlinear recurrence adds no clear gain. A supplementary test adds 5\,mm of noise to the visible puck positions. Without retraining, the $k{=}0$ student keeps a higher save rate than GRU-64 under this noise and 400\,ms blackout. The noise changes only how the puck is measured. The task and its dynamics are unchanged.

\begin{figure*}[t]
    \centering
    \includegraphics[width=\textwidth]{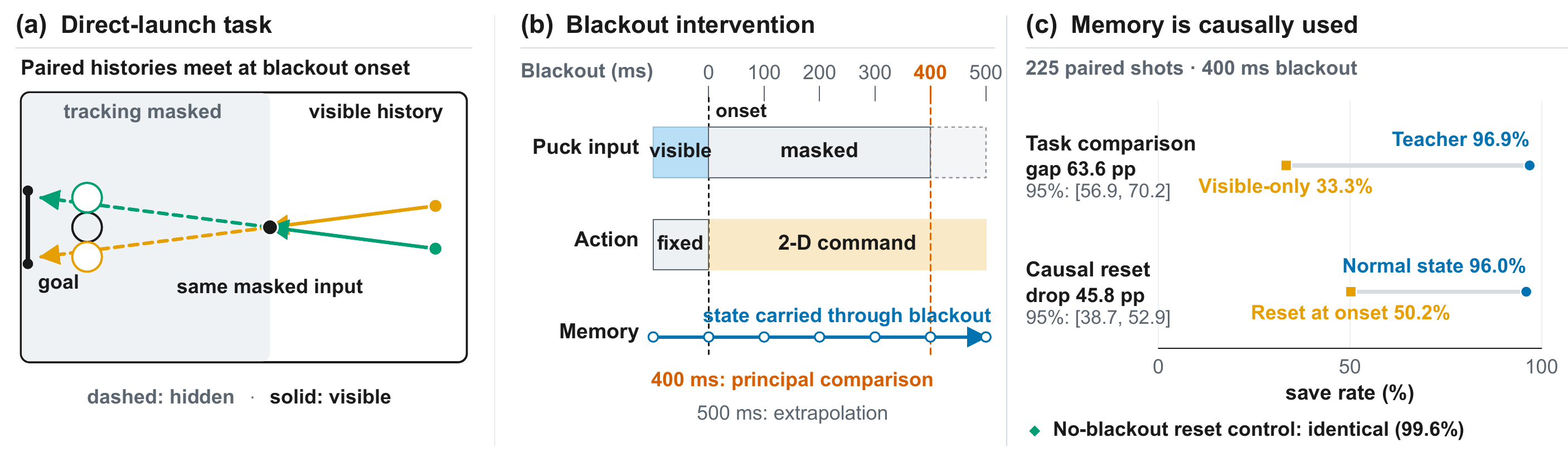}
    \caption{Why the task needs memory. (a) The two shots of an observation-alias pair reach the same point when tracking stops, then head for opposite sides of the goal; only the puck's earlier motion tells the defender which way to move. (b) Tracking stops after 100\,ms while the policy keeps acting; blackouts run up to 500\,ms, with 400\,ms the main comparison. (c) At 400\,ms the teacher clearly outperforms a policy without memory, and drops sharply when its state is erased at blackout onset. Erasure during visible tracking has no effect. Same 225 shots; paired bootstrap 95\% intervals.}
    \label{fig:task-memory}
\end{figure*}

\section{Related Work}

\paragraph{Robot air hockey.} In the robot air-hockey challenge~\cite{liu2024retrospective}, an agent controls a robot arm that plays air hockey against an opponent. The agent receives low-dimensional observations of the puck and the arm, and must return a command every 20\,ms, a control rate of 50\,Hz. Our teacher builds on the model-based agent developed for this challenge~\cite{orsula2024airhockey}. It is a DreamerV3 agent that outputs a target point for the mallet in the plane, which a fixed lower-level controller turns into joint commands. Deployments of challenge systems on real robots have met sensing dropouts, delays of tens of milliseconds, and strict timing constraints~\cite{liu2024retrospective}. 

Prior work treats these as deployment problems to engineer around. We instead isolate one of them, temporary loss of puck tracking, and make it the controlled variable of a one-shot defence task. This trades full-match realism for exact replay and paired evaluation.

\paragraph{Memory under partial observability.} When the puck disappears, the current observation is not enough and the policy needs memory. Recurrent networks are the classical approach to solve these problems. They keep performing even when observations randomly flicker off~\cite{hausknecht2015deep}, and real robots rely on the same idea, falling back on memory whenever their cameras fail in the field~\cite{miki2022learning}. World model agents such as DreamerV3 also carry memory in a recurrent latent state~\cite{hafner2023mastering}. What remains unclear is what kind of memory a given task needs. Across a broad benchmark, no memory architecture wins on every task~\cite{morad2023popgym}. That is the question we target. We use DreamerV3 only as a strong teacher that already has memory, and ask how small and how simple a recurrent student can be while still copying its behaviour when tracking is lost.

\paragraph{Behavioural policy transfer.} Copying a large teacher into a small student is a proven recipe. Policy distillation showed that a compact network can reproduce a much larger agent's behaviour~\cite{rusu2015policy}. A similar idea is widely used in robotics, where a teacher trained in simulation with access to rich information transfers its behaviour to a lightweight recurrent student that operates using only a history of sensor observations~\cite{lee2020learning}. We use this recipe because, with the teacher frozen, every student copies the same behaviour, so any difference between students can only come from their architecture. More specifically, we match the teacher's actions only~\cite{rusu2015policy} instead of its internal state. A student acting on its own can drift into situations that are poorly represented in the teacher-generated dataset. Small imitation errors then compound. Each unfamiliar state produces a larger error at the next step, and the gap can grow with the episode horizon. We therefore add one round of dataset aggregation~\cite{ross2011reduction}. We collect the states each student visits, ask the frozen teacher what it would have done there, and retrain on the combined data. To keep the comparison controlled, all student families are given the same aggregation budget.

\paragraph{Structured recurrence.} Finally, we consider how much recurrent nonlinearity the student actually needs. There are two existing lines of work that make opposite choice. Low-rank recurrent networks show that the rank of a nonlinear recurrent network controls what it can compute~\cite{mastrogiuseppe2018linking}. At the other end, linear recurrent models have shown that complex nonlinear recurrence is not always necessary. Structured linear models using diagonal and low-rank~\cite{gu2021efficiently} or diagonal~\cite{orvieto2023resurrecting} dynamics can effectively model long sequences. These ideas have also been extended to reinforcement learning under partial observability~\cite{lu2023structured}. Related work has also considered that trained nonlinear recurrent networks can be approximated by linear models~\cite{stolzenburg2025efficient}. What remains less explored between these two extremes, especially on control tasks where memory is essential, is a controlled comparison. We address this with a model that combines a diagonal linear transition with a rank-$k$ nonlinear component. By varying $k$ while keeping the teacher, dataset, training budget, and remaining network architecture fixed, we can isolate the effect of recurrent nonlinearity.

\section{Method}
\subsection{Teacher and policy boundary}
All teacher and student policies receive the same public interface. Its 19 values comprise seven joint positions, seven joint velocities, the planar positions of the mallet and puck, and a visibility flag. The action is a two-dimensional normalised mallet target, converted to low-level commands by the unchanged upstream controller with fixed stiffness and damping. The interface supplies no observation stack, opponent state, remaining blackout duration, puck velocity or privileged simulator state. The finite-history baseline explicitly buffers ten public puck observations; the other policies receive only the current observation at each step. Estimating puck motion therefore requires information carried across control steps, even while the puck is visible.

The teacher is a DreamerV3 agent~\cite{hafner2023mastering}, using the \texttt{size1m} preset. It was trained from scratch on the task reward for one million environment steps. We evaluated the retained checkpoints on 1\,125 paired validation episodes and selected the checkpoint at 700\,000 steps based on its validation save rate. The selected checkpoint is frozen and used without further parameter updates during data collection or evaluation.

For distillation, causal state ablation, principal evaluation and the noise extension, we run the teacher deterministically, taking the mode of its categorical latent state and the mean actor action. The earlier PPO memory comparison uses the upstream evaluation path. Recurrent state is reset only when an episode ends, for both teacher and recurrent students, except during the explicit state-ablation intervention. A visibility change otherwise does not reset the policy state. At each step, a recurrent student receives the current public observation together with its previously requested command. Both the recurrent state and previous command are initialised consistently at the beginning of each episode.

\subsection{Structured recurrent student}
We keep the student architecture fixed across variants and vary only the recurrent core. A two-layer sigmoid-weighted linear unit (\texttt{SiLU}) encoder maps the 19-dimensional public observation to a 32-dimensional representation $x_t$. The observation values are partly redundant, and 32 features are enough to compress them. The recurrent state $z_t$  has dimension $n=64$, larger than the encoder output to leave room for the bank of memory timescales described below, and equal to the nominal state size of the GRU baseline. For the action, the current features and recurrent state are concatenated and passed through a 64-unit \texttt{SiLU} hidden layer, followed by a two-dimensional $\tanh$ output for the mallet command. All widths were fixed before the principal runs and are identical across families, so any performance difference between variants can be attributed to the structure of the recurrent update rather than to the surrounding policy network.

The recurrent state is updated according to
    \[
    \begin{aligned}
    z_t={}&A z_{t-1}+B_xx_t+B_aa_{t-1}+b_z\\
          &+U\tanh(Vz_{t-1}+W_xx_t+W_aa_{t-1}+b_r)
    \end{aligned}
    \]
where $a_{t-1}$ is the command requested at the previous control step. The first line provides a simple linear memory mechanism. Each component of the previous state is propagated independently through the diagonal matrix $A$, with contributions from the current observation and previous action. The second line allows the model to modify this through a low-rank nonlinear pathway. The dimension $k$ of this pathway determines how many nonlinear combinations of recurrent state can directly influence the update.

We use
\[
A=\operatorname{diag}(\tanh\alpha)
\]
with learned $\alpha$. This guarantees that the diagonal entries and the eigenvalues of $A$ remain between $-1$ and $1$. This bound keeps the memory stable by construction. No learned setting of $\alpha$ can make the state grow on its own, so the recurrence cannot diverge during long rollouts, and training never has to trade stability for fit. The diagonal structure also keeps recurrent propagation linear in the state dimension, with a cost of $O(n)$ per control step. Individually, the 64 state channels behave as leaky memory traces with learned timescales. We initialise these timescales logarithmically between 40\,ms and 2\,s. Since the longest blackout lasts 500\,ms, this initialisation provides memory timescales both shorter and substantially longer than the period for which puck observations may be unavailable.

The innovation rank $k$ directly controls the amount of state-dependent nonlinearity in the recurrent update. In particular,
\[
\operatorname{rank}\left(
\frac{\partial z_t}{\partial z_{t-1}}-A
\right)\leq k.
\]
We verify this property by automatic differentiation for each variant. Importantly, this bound applies only to the nonlinear correction of the instantaneous recurrent Jacobian, and it does not restrict the recurrent state dimension or the complexity of the encoder and action head. When $k=0$, there is no innovation branch and the recurrent update is fully linear in the previous state, encoded observation, and previous action. We compare $k\in\{0,1,2,4\}$ while holding all remaining architectural dimensions fixed. Within each matched training seed, non-innovation parameters are initialised identically across ranks, so the variants differ at initialisation only through their innovation branch.

We control the nonlinear pathway by rank for two reasons. First, rank is exact and checkable. The rank of the state-to-state Jacobian correction can be verified by automatic differentiation, so the amount of nonlinear recurrence becomes a measured experimental variable. Second, rank connects the two literatures this study sits between: structured state-space models built on diagonal plus low-rank transitions~\cite{gu2021efficiently}, and low-rank recurrent networks, where rank controls what computations the network can perform~\cite{mastrogiuseppe2018linking}. One dial therefore runs from a fully linear update at $k{=}0$ towards unconstrained nonlinear recurrence.

\begin{figure*}[t]
    \centering
    \resizebox{0.98\textwidth}{!}{\definecolor{figblue}{HTML}{0072B2}
\definecolor{figsky}{HTML}{56B4E9}
\definecolor{figorange}{HTML}{824D00}
\definecolor{figgrey}{HTML}{5C6670}
\definecolor{figlight}{HTML}{EEF1F4}

\begin{tikzpicture}[
    >=Latex,
    font=\small,
    line cap=round,
    line join=round,
    input/.style={
        draw=black!75,
        fill=white,
        rounded corners=1.5pt,
        minimum width=2.25cm,
        minimum height=1.05cm,
        align=center,
        inner sep=5pt
    },
    block/.style={
        draw=black!75,
        fill=white,
        rounded corners=1.5pt,
        minimum height=1.18cm,
        align=center,
        inner sep=5pt
    },
    backbone/.style={
        block,
        draw=figblue,
        fill=figsky!10,
        minimum width=4.45cm,
        inner xsep=7pt
    },
    innovation/.style={
        block,
        draw=figorange,
        dashed,
        fill=figorange!9,
        minimum width=4.45cm,
        inner xsep=7pt
    },
    state/.style={
        draw=figblue,
        fill=figsky!18,
        circle,
        minimum size=1.25cm,
        align=center,
        inner sep=2pt
    },
    sum/.style={
        draw=black!75,
        fill=white,
        circle,
        minimum size=0.62cm,
        inner sep=0pt,
        font=\large
    },
    note/.style={
        align=left,
        font=\footnotesize,
        text=figgrey
    },
    flow/.style={-Latex, line width=1.0pt, draw=black!78},
    linearflow/.style={-Latex, line width=1.5pt, draw=figblue},
    innovationflow/.style={-Latex, line width=1.5pt, draw=figorange, dashed},
    branch/.style={line width=0.9pt, draw=black!65}
]

\node[input] (observation) at (0,1.75) {
    public observation\\[-1pt]
    $o_t\in\mathbb{R}^{19}$
};
\node[input] (carry) at (0,0) {
    carried values\\[-1pt]
    $z_{t-1}\in\mathbb{R}^{64}$, $a_{t-1}\in\mathbb{R}^{2}$
};
\node[block, minimum width=2.55cm] (encoder) at (3.45,1.75) {
    observation encoder\\[-1pt]
    $19\!\to\!64\!\to\!32$\\[-1pt]
    \footnotesize SiLU
};
\draw[flow] (observation) -- (encoder);

\node[block, fill=figlight, minimum width=2.20cm] (recurrentinputs) at (3.45,0) {
    recurrent inputs\\[-1pt]
    $e_t,\ z_{t-1},\ a_{t-1}$
};
\draw[flow] (encoder.south) -- (recurrentinputs.north);
\draw[flow] (carry.east) -- (recurrentinputs.west);

\node[backbone] (linear) at (8.1,0.92) {
    \textbf{diagonal linear backbone}\\[2pt]
    $\tanh(\alpha)\odot z_{t-1}+B_e e_t+B_a a_{t-1}+b_z$
};
\node[innovation] (innovation) at (8.1,-1.08) {
    \textbf{rank-$k$ nonlinear innovation}\\[2pt]
    $U\tanh\!\left(Vz_{t-1}+W_e e_t+W_a a_{t-1}+b_r\right)$
};
\node[sum] (sum) at (11.60,0.00) {$+$};
\node[state] (state) at (12.85,0.00) {$z_t$\\[-2pt]\scriptsize $64$};

\coordinate (corefork) at (4.95,0);
\draw[branch] (recurrentinputs.east) -- (corefork);
\fill[black!65] (corefork) circle (1.5pt);
\draw[branch] (corefork) |- (linear.west);
\draw[branch] (corefork) |- (innovation.west);
\fill[figblue] (linear.west) circle (1.8pt);
\fill[figorange] (innovation.west) circle (1.8pt);

\draw[linearflow] (linear.east) -- (sum.north west);
\draw[innovationflow] (innovation.east) -- (sum.south west);
\draw[flow] (sum) -- (state);

\node[block, minimum width=2.75cm] (head) at (15.28,0.00) {
    action head\\[-1pt]
    $[z_t,e_t]:96\!\to\!64\!\to\!2$\\[-1pt]
    \footnotesize SiLU, then tanh
};
\node[input, minimum width=1.55cm] (action) at (17.78,0.00) {
    action\\[-1pt]
    $a_t\in\mathbb{R}^{2}$
};
\draw[flow] (state) -- (head);
\draw[flow] (head) -- (action);

\draw[flow]
    (encoder.north east)
    -- ++(0,0.92)
    -| node[pos=0.58, above=3pt, fill=white, inner sep=1.5pt,
       font=\footnotesize, text=figgrey]
       {$e_t$ to action head}
    (head.north);

\node[note, anchor=north] at (8.45,-1.82) {
    \textcolor{figorange}{$k=0$: dashed branch absent}\\
    \textcolor{figorange}{$k\in\{1,2,4\}$: $64\!\to\!k\!\to\!64$, rank at most $k$}
};

\end{tikzpicture}}
    \caption{The structured student family: a shared diagonal linear backbone
(solid blue) plus an optional rank-$k$ nonlinear innovation (dashed orange,
absent for $k{=}0$). The encoder and action head stay nonlinear in every
variant.}
    \label{fig:structured-student}
\end{figure*}

\subsection{Behavioural policy distillation}
Students are trained to imitate the frozen teacher using behavioural distillation. The target at each step is the teacher's deterministic action mean after public action clipping, and the student is optimised using mean squared error between its command and the teacher target. Only teacher actions are used as supervision. Teacher hidden states, latent variables, and true puck state are not exposed to the student. All student families are initially trained on the same teacher-generated dataset of 20\,000 complete episodes, containing 713\,257 transitions.

Training only on teacher-controlled trajectories introduces a distribution shift at evaluation time, since student actions may lead to states that are not represented in the original dataset. To reduce this effect, we apply one round of DAgger-style data collection for each student family~\cite{ross2011reduction}. The five matched seeds in each family are first trained on the common teacher dataset. Each collector then runs 4\,000 closed-loop episodes under the same paired shot and blackout schedule, which generates 20\,000 shadow episodes for each family.

The frozen teacher then relabels each valid step along these student-generated trajectories. During relabelling, the teacher maintains its own recurrent state while following the observation sequence generated by the student. At each step, it receives the same public observation available to the student together with the student's previously requested command, and provides the action it would have taken along that trajectory. Shadow trajectories are collected and retained separately for each student family.

Final training combines each family's 20\,000 shadow episodes with the 20\,000 common teacher episodes. We then train five new models from scratch rather than continuing from the collector checkpoints for each family. The optimisation budget is identical across families, and no family-specific hyperparameter tuning is performed. All families share one optimisation recipe, fixed before principal runs. We train with AdamW at learning rate $3{\times}10^{-4}$ and weight decay $10^{-5}$, in batches of 128 episodes, for at most 100 epochs with early stopping on validation action error, and clip gradient norms at 1.0. Sequences are trained in 64-step chunks, half the 128-step episode cap, so an episode spans at most two chucks while the depth of backpropagation stays bounded. Checkpoint selection uses only offline validation action error and is independent of closed-loop performance.

Recurrent training uses sequences of 64 steps. When an episode spans multiple training chunks, its recurrent state is carried from one chunk to the next rather than reset. The imitation loss is evaluated at every valid step and weighted so that each complete episode contributes equally regardless of duration. Finally, although collection schedules and episode budgets are identical across families, episode length depends on the behaviour policy. Realised transition and teacher-query counts can therefore differ, and are reported separately for each family.

\subsection{Comparisons and measurements}
We compare seven distilled student families with different forms of memory. Four students use the structured recurrent architecture with a fixed state dimension of $n=64$ and innovation rank $k\in\{0,1,2,4\}$. A GRU-64 uses the same encoder, action head, 64-dimensional recurrent state and previous-command input, but replaces the structured recurrent update with a conventional GRU.

We also include two non-recurrent alternatives. The finite-history student uses a multilayer perceptron (MLP) over current proprioception and a ten-step history of masked puck observations (approximately 200\,ms). The memoryless student receives only the current observation. In addition, we train a separate memoryless PPO policy with two 256-unit \texttt{tanh} hidden layers directly from reward. This policy is used only as a task-validity control and is not included among the distilled families.

All distilled families share the same teacher dataset, shadow data collection schedule and episode budget, optimisation budget, five matched training seeds, paired shot and blackout schedule, and inference protocol. Shadow trajectories are family-specific since they are generated by each family's own collector policies. This is the distribution shift that the DAgger-style correction is intended to capture.

We evaluate behaviour primarily through save rate across blackout durations, and additionally report the gap relative to the teacher, per-step action error against teacher labels, and contact-aware outcome categories to help interpret differences in closed-loop behaviour. Efficiency is measured separately in terms of storage, computation and runtime. Storage includes total and recurrent-core parameter counts, together with carried-state and recurrent-state memory. Computation is reported as multiply-adds per step, and runtime is measured using batch-one CPU latency. We report these separately because parameter count, arithmetic cost and measured execution time do not need to scale together.

The main evaluation uses a fresh set of 225 shots at blackout lengths $\{0,5,10,15,20,25\}$ steps, with 25 steps retained as the predeclared extrapolation condition, giving 1\,350 episodes per checkpoint and 6\,750 per family. All 35 final checkpoints, paired validation results, and efficiency measurements were frozen before the test split was opened. The split was released for a single final evaluation, with no subsequent retraining or checkpoint reselection.

Family-level uncertainty is estimated with a 10\,000-replicate paired hierarchical cluster bootstrap over training seeds and evaluation units, using common resamples across compared families. Observation-alias pairs are treated as single clusters so that their two shots remain paired. The single-policy comparisons do not contain a training-seed level and therefore use a paired episode bootstrap over matched shot-blackout conditions.

For the post-hoc subgroup analysis and supplementary noise experiment, we use 10\,000 paired bootstrap replicates over training seeds and shot units. Each alias pair is one unit; individual support shots are stratified by target region. Resampling preserves the observed stratum sizes and uses common draws across policies and, for the noise experiment, noise levels and blackout durations. The teacher has no training-seed resampling. These percentile intervals are pointwise 95\% intervals without multiplicity adjustment and do not replace the principal intervals.

Latency measurements use batch-one float32 inference on an Intel Core Ultra 9 285 CPU (24 cores), with one process and one thread pinned to logical CPU 0 under Linux. Each checkpoint receives 10\,000 warm-up calls followed by ten blocks of 10\,000 timed calls. We compute the median and 95th percentile of the ten block-mean latencies, then report the median of each statistic across five training seeds. These are not percentiles of individual-call latency. Timing includes the encoder, memory update and action head, but excludes simulation, rendering, inverse kinematics and checkpoint loading. Structured models use a compiled float32 reduction kernel verified bit-exact against the canonical evaluator; GRU and non-recurrent baselines use BLAS-backed NumPy. We therefore treat multiply-adds as the primary architecture-level cost measure and latency as implementation-specific.

\paragraph{Reproducibility.} Code, configurations, analysis and figure-generation scripts are available in the public repository\footnote{\raggedright\href{https://github.com/unswei/airhockey-distillation}{\nolinkurl{github.com/unswei/airhockey-distillation}}} under the tag \href{https://github.com/unswei/airhockey-distillation/tree/acra-2026-submission}{\texttt{acra-2026-submission}}.

\section{Air-Hockey Tracking-Loss Task}
We study a controlled air-hockey set-piece in MuJoCo. A simulated KUKA iiwa defends its goal against a directly launched puck, at 50\,Hz with a 2.5\,s timeout. Launching the puck directly, rather than through an opponent, lets us replay the same shot exactly across policies and blackout conditions. The policy maps the current public observation, plus any internal state it keeps, to a mallet target that a fixed low-level controller executes. The simulator is deterministic, puck velocity is never observed, and each episode contains one blackout whose length the policy does not know. This isolates tracking loss, at the cost of representing a set-piece rather than a complete play.

Each episode opens with a 100\,ms visible prefix (five control steps) during which the mallet is held fixed, so the pre-blackout history cannot leak into the robot's own motion. The policy takes control at the next step, when masking begins. The puck is then masked for one contiguous interval of $\{0,5,10,15,20\}$ steps (0--400\,ms), with a predeclared 25-step (500\,ms) condition to test extrapolation. While masked, the puck position and visibility flag read zero. Observations are exact in the principal experiments. Noise is only added in the supplementary test.

To make memory necessary, we construct \emph{observation-alias} pairs. The two shots of a pair reach nearly the same puck position when masking begins, having approached along different trajectories, and continue towards opposite sides of the goal. Because the mallet is still fixed at that moment, masking produces identical public observations, and only the preceding observation history separates the two shots. A deterministic memoryless policy must therefore respond identically to both members of a pair. The guarantee holds at blackout onset. Once control begins, observations may diverge as policies act.

We audit the generated trajectories to confirm this ambiguity in the realised simulator states. Across 90 alias families, paired shots differ by at most 8.1\,mm in their last visible puck position and produce bit-identical public observations at blackout onset, where a privileged controller requires actions at least 0.315 apart in the normalised action space. The audit establishes ambiguity at the start of blackout not identical sequences throughout the defence.

The \texttt{direct\_launch\_v3} distribution contains 900 training, 225 validation, and 225 task-validation shots generated from independent split seeds. The principal test uses a separate 225-shot split held out until the evidence release. Returns, arrests and safe deflections count as saves; all other outcomes, including simulator or safety faults, are scored as failures. The training reward 
($+1$ save, $-1$ concession, $+0.2$ first contact) is used only for the reward-trained teacher and PPO baseline.

Calibration runs on 216 further shots bracket the difficulty. An inactive defender concedes 208 shots and a fixed-centre defender concedes 209, while a privileged controller with the true puck state saves 214, with no simulator faults. The task is hard without a real defensive response and almost fully solvable when the puck state is available.

\paragraph{Supplementary noise setup.}\label{sec:noise-setup}
To test sensitivity to measurement error, we evaluate the frozen teacher and all five existing seeds of $k=0$, $k=4$ and GRU-64 without retraining. These students compare the linear update with the largest tested innovation rank and a conventional recurrent baseline. We use 225 fresh shots from the same distribution (90 alias pairs and 45 support shots), crossed with 0 and 400\,ms blackouts and per-coordinate noise standard deviations of 0, 1 and 5\,mm: 21\,600 episodes. This extension was designed after inspecting the principal results; its conditions and analysis were fixed before evaluation.

At every visible observation, including the action-locked prefix, we add zero-mean Gaussian noise independently across coordinates and time to the puck position, converting millimetres to normalised coordinates before clipping. Each shot unit has one standard-normal trace, scaled across noise levels and shared across policies, training seeds and blackout durations; alias-pair members share a trace. Hidden positions remain zero, and visibility, proprioception, dynamics and save criteria are unchanged. The clean condition uses these same fresh shots, not the principal-test rates. The noise levels are synthetic, not calibrated to a physical tracker.

\section{Results}
\subsection{Memory requirement}
We first test whether successful defence under tracking loss requires memory. Using the same 225 shots at each blackout length (1\,125 paired episodes per policy), we compare the teacher with the best validation-qualified seed from three independently trained memoryless PPO policies. With no blackout, both perform well. The teacher saves 99.6\% of shots and PPO saves 90.7\%. However, at 400\,ms, the teacher retains a 96.9\% save rate while PPO falls to 33.3\%. This gives a difference of 63.6 percentage points, with a 95\% confidence interval of $[56.9, 70.2]$. The gap increases by 54.7 points from 0 to 400\,ms (Fig.~\ref{fig:task-memory}c), showing that the performance difference grows specifically with tracking loss.

To test whether the teacher relies on its recurrent memory, we reset its recurrent state at the onset of each blackout. No other aspect of inference or evaluation is changed. Under a 400\,ms blackout, the reset reduces the teacher's save rate from 96.0\% (the paired baseline of this separate run) to 50.2\%, a paired decrease of 45.8 percentage points, with a 95\% confidence interval of $[38.7, 52.9]$. In the non-blackout condition, the reset is never triggered and all 225 episode records remain byte-identical between runs. The teacher's robustness to blackout therefore depends strongly on information retained in its recurrent state.

\subsection{Closed-loop imitation and distribution correction}
Offline imitation alone did not guarantee reliable closed-loop behaviour. An initial diagnostic found that, at the 95th percentile, the nearest-history distance from student-generated trajectories to the training set was 16.8 times larger than for teacher-controlled validation trajectories. After an initial sequence-loss weighting error was corrected, a $k=2$ pilot trained only on teacher-controlled trajectories still saved only 67.1\% of non-blackout validation shots. The distance diagnostic and corrected pilot are separate runs.

We therefore apply the shadow-teacher correction. After one correction round, the same $k=2$ architecture and seed achieved a 98.7\% save rate without blackout and 98.4\% across the paired validation grid. The experiment is intended to diagnose distribution shift rather than to establish an advantage specific to $k=2$. Every student family in the main study receives the same correction schedule and collection budget.

\subsection{Principal recurrent-complexity comparison}
\begin{figure*}[!t]
    \centering
    \begin{minipage}[t]{0.46\textwidth}
        \centering
        \includegraphics[width=\linewidth]{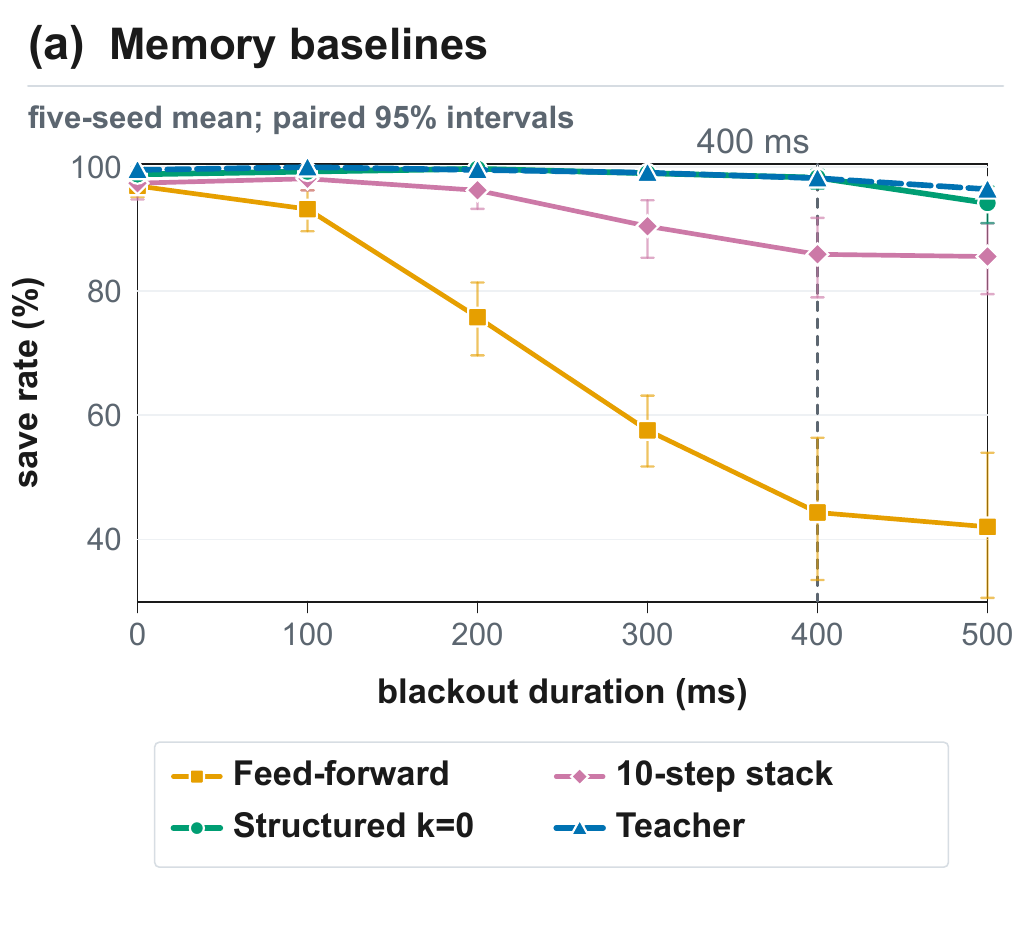}
    \end{minipage}\hfill
    \begin{minipage}[t]{0.46\textwidth}
        \centering
        \includegraphics[width=\linewidth]{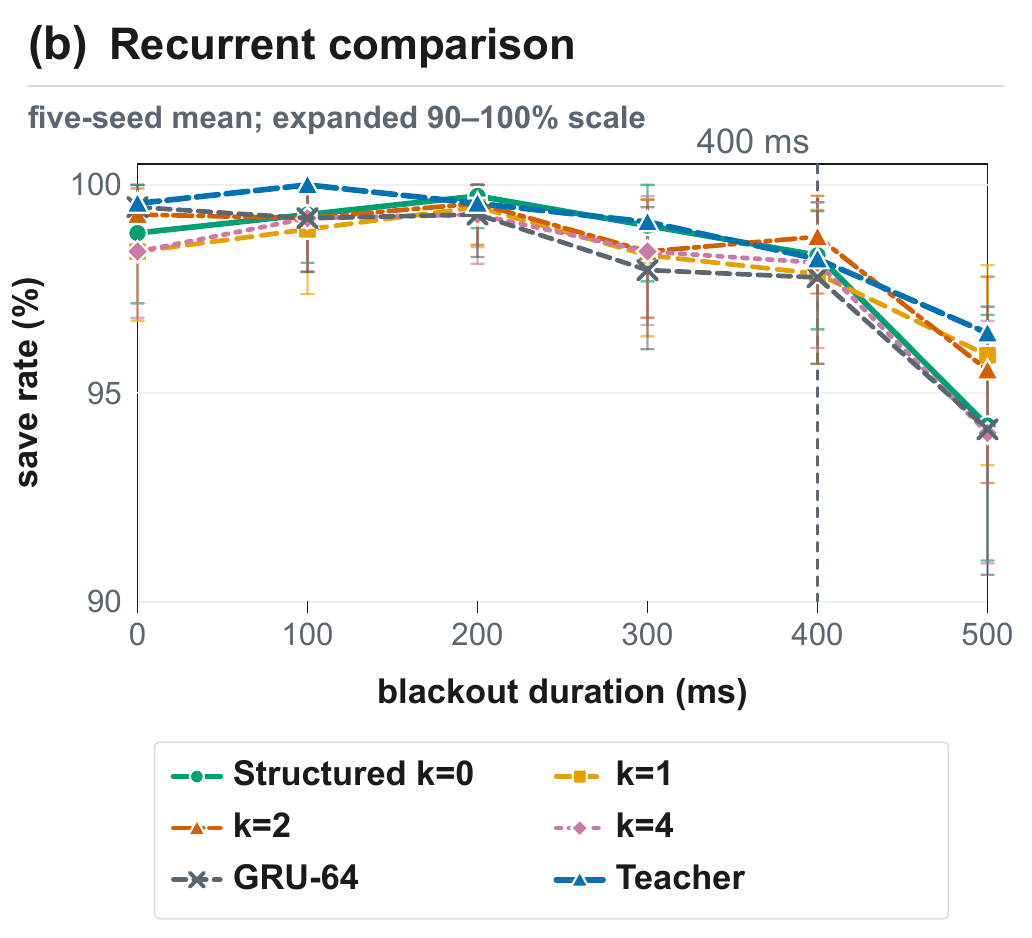}
    \end{minipage}
    \par\medskip
    \includegraphics[width=0.9\textwidth]{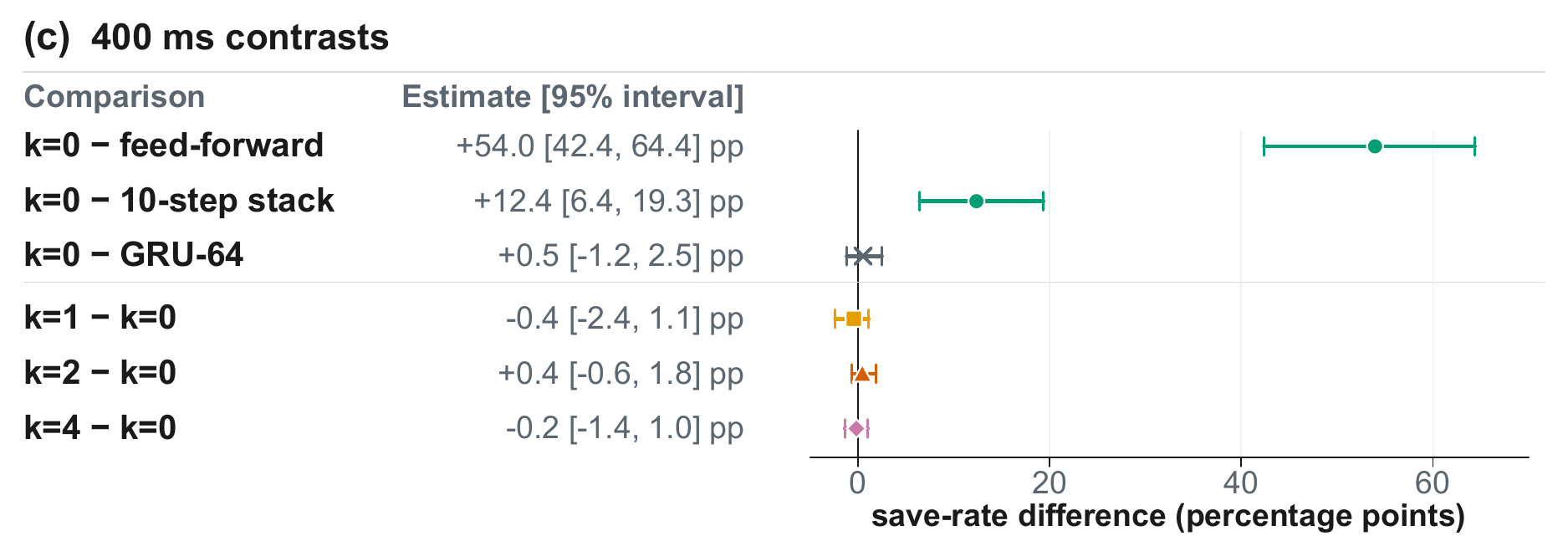}
    \caption{Principal behavioural results. (a) The feed-forward student deteriorates sharply once the puck is hidden; a ten-step observation stack helps, while the structured $k{=}0$ student stays close to the teacher through the main 400\,ms range. (b) On a 90--100\% scale, all four structured students and GRU-64 stay close to the teacher; the 500\,ms point extrapolates beyond the main range. (c) Paired differences at 400\,ms, positive favouring the first policy named: $k{=}0$ clearly beats feed-forward and finite-stack memory, with no clear difference from GRU-64 or from rank-1, 2, or 4 corrections. Student curves are means over five matched seeds against one frozen teacher; bars and intervals are paired bootstrap 95\%.}
    \label{fig:principal-behaviour}
\end{figure*}

\begin{table*}[t]
\centering
{\footnotesize
\setlength{\tabcolsep}{3.0pt}
\renewcommand{\arraystretch}{1.12}
\begin{tabular}{@{}lrrrrrrrr@{}}
\toprule
Family &
\shortstack{400 ms save\\(\%, 95\% CI)} &
\shortstack{Overall\\save (\%)} &
\shortstack{Total\\params} &
\shortstack{Recurrent\\core params} &
\shortstack{Carried\\state (B)} &
\shortstack{Recurrent\\state (B)} &
\shortstack{MACs\\per step} &
\shortstack{CPU latency\\median / p95 ($\mu$s)} \\
\midrule
Feed-forward       & 44.4 $[33.5,56.4]$ & 73.6 & 5,602  & 0      & 0   & 0   & 5,440  & 12.97 / 13.08 \\
Ten-step stack     & 86.0 $[79.0,91.8]$ & 93.7 & 7,330  & 0      & 120 & 0   & 7,168  & 17.86 / 17.99 \\
Structured $k=0$  & 98.3 $[96.5,99.6]$ & 99.0 & 12,002 & 2,304  & 264 & 256 & 11,776 & 23.03 / 23.15 \\
Structured $k=1$  & 97.9 $[95.7,99.4]$ & 98.6 & 12,165 & 2,467  & 264 & 256 & 11,938 & 28.20 / 28.39 \\
Structured $k=2$  & 98.8 $[97.4,99.7]$ & 99.0 & 12,328 & 2,630  & 264 & 256 & 12,100 & 28.26 / 28.43 \\
Structured $k=4$  & 98.1 $[96.1,99.6]$ & 98.7 & 12,654 & 2,956  & 264 & 256 & 12,424 & 28.31 / 28.61 \\
GRU-64             & 97.8 $[95.7,99.4]$ & 98.7 & 28,898 & 19,200 & 264 & 256 & 28,352 & 40.06 / 40.33 \\
\bottomrule
\end{tabular}
}
\caption{Complete comparison of the seven student families. The 400 ms column
reports the five-seed mean and paired-bootstrap 95\% confidence interval.
Overall save rate pools the five predeclared 0--400 ms conditions and excludes
the 500 ms extrapolation. For each latency statistic (median or p95), entries are medians across five
seeds of that statistic computed from block-mean latencies on one pinned
Intel Core Ultra 9 285 CPU core. Carry includes every value retained between control steps; recurrent
state excludes the previous action and the finite observation stack.}
\label{tab:complete-comparison}
\end{table*}

Table~\ref{tab:complete-comparison} and Fig.~\ref{fig:principal-behaviour} summarise the main test. At a 400\,ms blackout, the feed-forward student saves 44.4\% of shots and the ten-step history student 86.0\%. Every recurrent student stays close to the teacher's 98.2\%, with save rates between 97.8\% and 98.8\%. Per-family intervals are given in Table~\ref{tab:complete-comparison}.

Memory type matters far more than memory complexity. The $k{=}0$ student outperforms the feed-forward student by 54.0 percentage points and the ten-step history student by 12.4 points. The stack holds up for shorter blackouts but falls behind once the blackout outlasts its roughly 200\,ms window. Both contrasts are clearly separated from zero (Fig~\ref{fig:principal-behaviour}c). Between recurrent architectures the differences are small. The paired $k{=}0$ minus GRU-64 difference is 0.5 percentage points with a 95\% interval of $[-1.2, 2.5]$, although the GRU recurrent core has 8.3 times as many parameters. Raising the innovation rank gives no consistent improvement either. All three paired differences from $k{=}0$ lie within 0.4 points of zero, and every interval includes zero (Fig.~\ref{fig:principal-behaviour}c).

At the predeclared 500\,ms extrapolation, recurrent students save between 94.0\% and 95.9\%, against 96.4\% for the teacher, 85.6\% for the ten-step student, and 42.0\% for the feed-forward student. Recurrent students remain ahead, but these rates do not establish equivalence among the recurrent architectures.

\begin{figure*}[t]
    \centering
    \begin{minipage}[t]{0.493\textwidth}
        \centering
        \includegraphics[width=\linewidth]{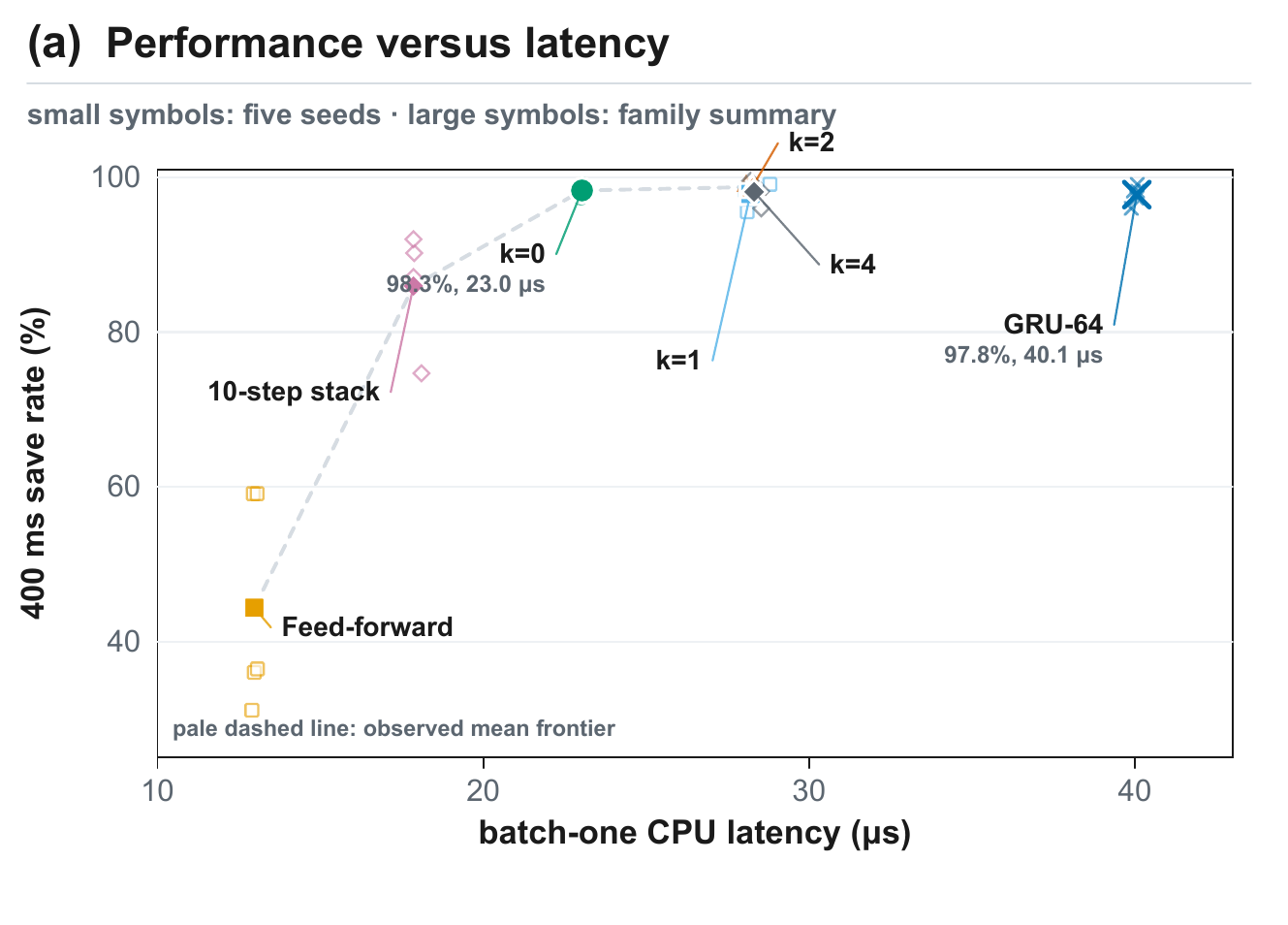}
    \end{minipage}\hfill
    \begin{minipage}[t]{0.493\textwidth}
        \centering
        \includegraphics[width=\linewidth]{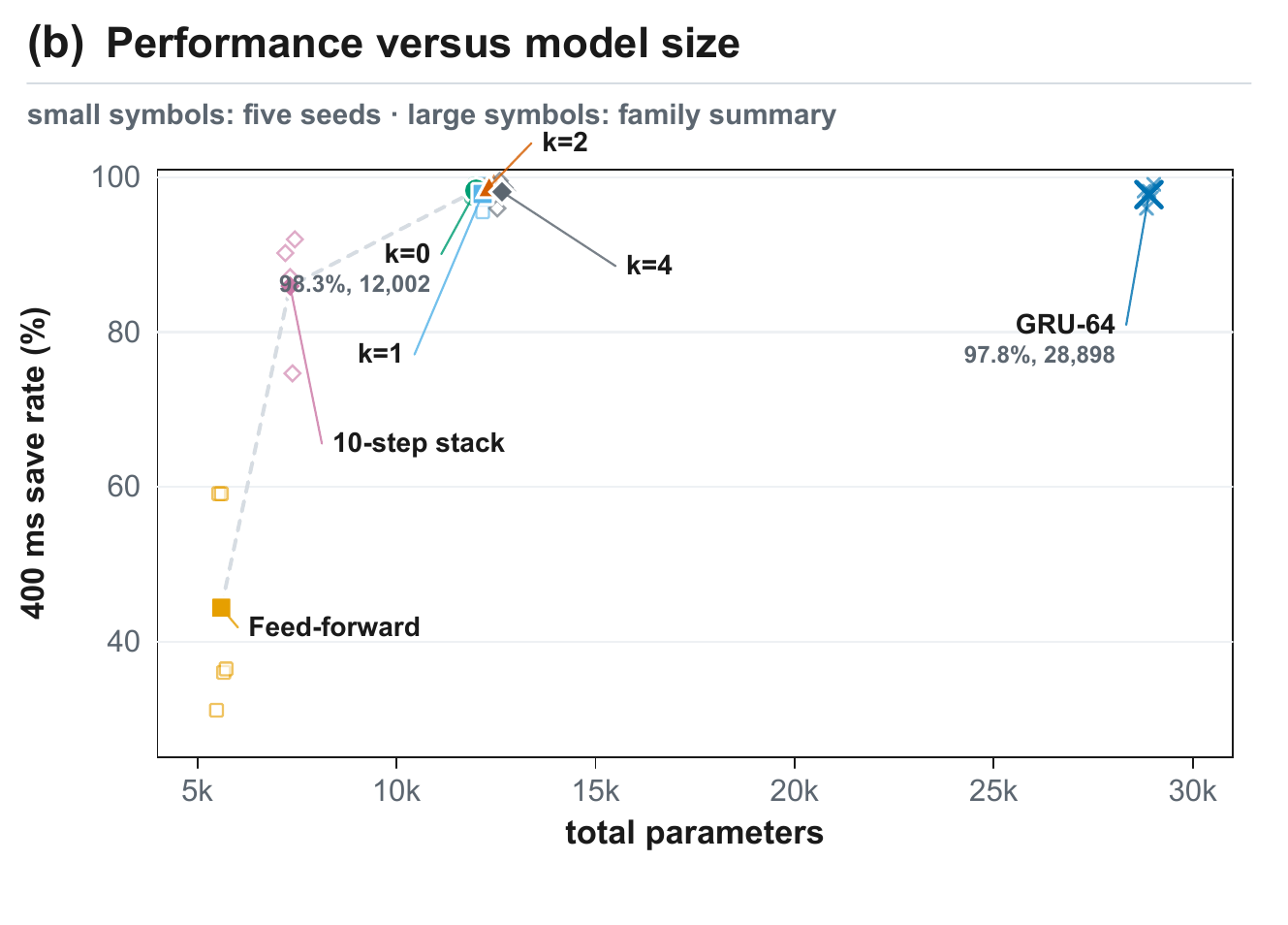}
    \end{minipage}
    \caption{Performance--cost frontier at 400\,ms tracking loss. Small symbols are the five matched seeds; large labelled symbols are family mean save rates, placed at the median measured latency in (a) and the parameter count in (b), where seed points are offset slightly because the count is fixed within a family. The dashed line joins the non-dominated families. Structured $k{=}0$ reaches 98.3\% saves with 12{,}002 parameters and 23.0\,$\mu$s latency; GRU-64 reaches 97.8\% with 28{,}898 parameters and 40.1\,$\mu$s. Their paired save-rate difference is 0.5 points, 95\% interval $[-1.2,2.5]$.}
    \label{fig:performance-cost-frontier}
\end{figure*}

\paragraph{Post-hoc shot breakdown.} We split the existing test records into 180 alias shots (90 pairs) and 45 support shots. Every structured model saves every alias shot, across all six blackout durations and five seeds, while the feed-forward student saves 50.9\% of them at 400\,ms. Alias shots therefore expose the memory contrast but cannot separate the structured models. Support shots are harder, and the comparison there is underpowered. At 400\,ms, $k{=}0$ and GRU-64 both save 91.6\%. At 500\,ms they fall to 71.1\% and 73.8\%, against 82.2\% for the teacher, and their paired difference is $-2.7$ points with a wide interval of $[-15.6, 10.7]$. The corresponding support rates for $k{=}1$, $k{=}2$ and $k{=}4$ are 79.6\%, 77.8\% and 70.2\%. These exploratory estimates leave room for differences among recurrent models, and the perfect alias rates are not guarantees for unseen shots.

\subsection{Efficiency and compression}
All recurrent students carry the same 64-dimensional float32 state, 256 bytes, or 264 with the previous command. The families differ in what it costs to update that state. The $k{=}0$ model has 12\,002 parameters against 28\,898 for GRU-64 (58.5\% fewer), a recurrent core of 2\,304 parameters against 19\,200 (88.0\% fewer), and needs 11\,776 rather than 28\,352 multiply-adds per step.

The CPU benchmark points the same way, with batch-one latency falling from 40.06\,$\mu$s for GRU-64 to 23.03\,$\mu$s for $k{=}0$. The two models use different implementations, so multiply-adds are the cleaner architecture-level comparison. The compiled structured kernel was verified against the canonical evaluator and reproduces the same validation trajectories exactly.

Figure~\ref{fig:performance-cost-frontier} shows that $k{=}0$ reaches a high save rate at the lowest measured cost among the recurrent families. The $k{=}2$ student has a slightly higher observed mean at additional cost, and GRU-64 reaches a similar save rate with more than twice the parameters and multiply-adds per step. The paired save-rate contrasts for both comparisons were reported above, and neither separates from zero.

\subsection{Sensitivity to puck-position noise}

Does noise in the visible puck positions change the recurrent comparison? Table~\ref{tab:observation-noise} reports every condition of the supplementary setup. At 5\,mm noise and a 400\,ms blackout, $k{=}0$ saves 98.5\% and GRU-64 92.9\%, a paired difference of 5.6 percentage points ($[2.8, 9.0]$). The $k{=}0$ minus $k{=}4$ difference is 4.0 points ($[1.1, 8.1]$). Relative to clean observations on the same shots, $k{=}0$ changes by $+0.4$ points ($[0.0, 1.1]$) while GRU-64 drops by 3.9 points ($[-6.9, -1.3]$).

\begin{table}[!ht]
\centering
\resizebox{\columnwidth}{!}{%
\begin{tabular}{@{}lrrrrrr@{}}
\toprule
 & \multicolumn{3}{c}{No blackout} & \multicolumn{3}{c}{400\,ms blackout} \\
\cmidrule(lr){2-4}\cmidrule(l){5-7}
Policy / $\sigma$ & 0 & 1 & 5 & 0 & 1 & 5 \\
\midrule
$k=0$ & 98.6 & 98.3 & 98.4 & 98.1 & 98.0 & 98.5 \\
 & \scriptsize$[97.2,99.6]$ & \scriptsize$[96.8,99.6]$ & \scriptsize$[97.1,99.6]$ & \scriptsize$[96.4,99.6]$ & \scriptsize$[96.4,99.5]$ & \scriptsize$[96.9,99.7]$ \\
\addlinespace[1.5pt]
$k=4$ & 98.5 & 98.6 & 98.9 & 97.6 & 97.5 & 94.5 \\
 & \scriptsize$[97.0,99.7]$ & \scriptsize$[97.2,99.7]$ & \scriptsize$[97.8,99.8]$ & \scriptsize$[95.8,99.1]$ & \scriptsize$[95.5,99.1]$ & \scriptsize$[89.9,98.0]$ \\
\addlinespace[1.5pt]
GRU-64 & 99.4 & 99.2 & 97.9 & 96.8 & 96.8 & 92.9 \\
 & \scriptsize$[98.3,100]$ & \scriptsize$[98.0,100]$ & \scriptsize$[95.7,99.5]$ & \scriptsize$[94.3,98.8]$ & \scriptsize$[94.3,98.8]$ & \scriptsize$[88.9,96.1]$ \\
\addlinespace[1.5pt]
Teacher & 99.1 & 99.1 & 98.7 & 97.3 & 96.9 & 96.0 \\
 & \scriptsize$[97.8,100]$ & \scriptsize$[97.8,100]$ & \scriptsize$[97.3,100]$ & \scriptsize$[95.1,99.1]$ & \scriptsize$[94.7,98.7]$ & \scriptsize$[93.3,98.2]$ \\
\bottomrule
\end{tabular}}
\caption{Save rates (\%) on 225 fresh shots per condition, without
retraining. Student entries average five seeds; the teacher is one frozen
policy. Noise $\sigma$ is per-coordinate Gaussian standard deviation in mm.
Brackets are paired pointwise 95\% intervals from the same cluster
bootstrap as the in-text contrasts.}
\label{tab:observation-noise}
\end{table}

The difference is clearest on alias shots, where $k{=}0$ saves all of
them at 5\,mm and 400\,ms against 94.1\% for GRU-64. The support-shot
difference stays unresolved (4.4 points, $[-1.3, 11.6]$). The ordering is
also not uniform. Without blackout at 5\,mm, $k{=}4$ saves more support
shots than $k{=}0$ ($-4.0$ points, $[-8.4, -0.9]$). The aggregate result
at 400\,ms therefore does not establish that the linear update is
preferable in every condition.

\section{Discussion and Limitations}
Memory is important for this task, but the recurrent update can be relatively simple. The memoryless baseline performs well with visible observations but degrades as blackouts lengthen. Resetting the teacher's state at blackout onset also reduces its save rate, showing that information retained before blackout contributes to defence. After correcting distribution shift under student control, recurrent students recover aggregate performance close to the teacher through 400\,ms.

In the clean-observation principal test at 400\,ms, the paired comparisons show no clear save-rate gain from increasing innovation rank or replacing the linear update with GRU-64. The noise extension shows that $k=0$'s high save rate does not depend on exact visible positions under the tested perturbations. It does not identify why its performance changes less than that of GRU-64 at 5\,mm and 400\,ms. The $k=0$ policy also remains nonlinear in its encoder and action head; only the recurrent update is linear.

Hidden puck motion is approximately ballistic over the short blackout, so preserving information from earlier positions may matter more than state-dependent nonlinear computation. Nonlinear processing remains available in the encoder and action head. This is a possible explanation of the clean-observation result, not a mechanism established by the experiments or an explanation of the noise difference.

The ten-step stack loses access to visible puck observations once a blackout exceeds its approximately 200\,ms window. Yet at 400\,ms it saves 94.6\% of alias shots and 51.6\% of support shots. The breakdown does not identify whether current robot state, earlier actions or other correlations explain its success after visible samples leave the window.

Distillation also exposes distribution shift: student actions lead to histories that differ from the teacher-controlled training data. Even after correcting the offline training procedure, imitation alone gives poor closed-loop performance. For the $k=2$ pilot, one round of shadow-teacher correction using teacher-labelled student trajectories largely closes this gap.

Student training targets the teacher's actions, not its latent state. Similar aggregate save rates in the principal test do not establish formal equivalence, which the analysis was not designed to test.

The experiments use one simulated air-hockey set-piece, one teacher, a 64-dimensional principal state and five training seeds per student family. Ceiling effects limit the clean comparison: all structured models save the 180 alias shots, while support-shot uncertainty remains broad at 500\,ms. The noise extension changes the observations but not the dynamics. One Gaussian trace per shot unit does not measure the full variation across repeated noise realisations, and the levels are not sensor-calibrated. Pointwise intervals across multiple conditions do not establish a general architecture ranking. Correlated or biased tracking errors, longer or repeated occlusions and physical robot deployment remain untested.

\section{Conclusion}
We asked whether a policy needs nonlinear recurrence to defend an air-hockey goal through temporary tracking loss, or whether a linear memory update is enough. Distilled from the same frozen teacher under a fixed protocol, the fully linear $k{=}0$ student matched every nonlinear alternative. Its 400\,ms save rate sits within 0.1 percentage points of the teacher and 0.5 points above GRU-64, and raising the innovation rank changes the paired outcome by less than half a point. Under 5\,mm observation noise, again without retraining, the linear student holds 98.5\% while GRU-64 falls to 92.9\%. It does this with 88.0\% fewer recurrent-core parameters and 58.5\% fewer multiply-adds per step at the same state size.

These results come from one controlled setting. They do not prove that the linear and nonlinear students are equivalent, or that linear memory is enough in general, and the support-shot comparison at 500\,ms stays unresolved. The results still carry a practical lesson. The puck moves predictably while hidden, and a linear memory was enough to track it. We expect the same on other tasks where the hidden quantity keeps evolving predictably, and we expect nonlinear recurrence to become necessary once the hidden part of a task can change behaviour while unseen, through contacts, mode switches or an opponent's decisions. We plan three follow-ups. First, a set of tasks whose hidden dynamics become gradually more complex, to find where linear memory stops being enough. Second, a benchmark built from the observation-alias construction, where the memory required by a task can be proven rather than assumed. Third, tests of the present result under correlated tracking errors and on a physical robot.

\paragraph{AI-use disclosure.}
The authors used OpenAI Codex for language
editing, code assistance, and workflow automation. The authors reviewed all
generated material and take responsibility for the ideas, claims, experiments,
results, and text.

\bibliographystyle{named}
\bibliography{references}

\end{document}